# A Review of Machine Learning based Anomaly Detection Techniques

Harjinder Kaur
Dept of Computer Science
and Engineering,
Punjabi University Regional
Centre of IT & Management,
Mohali, Punjab, India

Gurpreet Singh
Dept of Computer Science
and Engineering,
DAV Institute of Engineering
& Technology,
Jalandhar, Punjab, India

Jaspreet Minhas
Dept of Computer Science
and Engineering,
DAV Institute of Engineering
and Technology,
Jalandhar, Punjab, India

**Abstract**: Intrusion detection is so much popular since the last two decades where intrusion is attempted to break into or misuse the system. It is mainly of two types based on the intrusions, first is Misuse or signature based detection and the other is Anomaly detection. In this paper Machine learning based methods which are one of the types of Anomaly detection techniques is discussed.

## 1. INTRODUCTION
Intruders may be from outside the host or the network or legitimate users of the network. Intrusion detection is the process of monitoring the events that are occurring in the systems or networks and analyzing them for signs of possible incidents, which are violations or threats to computer security policies, acceptable use policies, or standard security practices [1]. That system which detects the intrusion in the system is known as IDS (Intrusion detection System).This concept has been around for two decades but recently seen a dramatic rise in the popularity and incorporation into the overall information security infrastructure [1]. Thus the intrusion can be found mainly using two classification techniques: Misuse or signature based detection and the other is Anomaly detection.

## 2. DETECTION TECHNIQUES
First technique of detection, Signature based also referred to as pattern based, looks for evidence known to be indicative of misuse. Whether it's looking for specific log entries or a specific payload in a data packet, the NIDS/HIDS is looking for something it knows about – a signature of misuse. While Anomaly based detection looks for signs that something is out of the ordinary that could indicate some form of misuse. Anomaly based systems analyze current activity against a "baseline" of "normal" activity and look for deviations outside that which is considered normal [2]. These two techniques applied for the major classification of IDS named Host based IDS (HIDS), Network based IDS (NIDS) and the Hybrid IDS [1].

## 3. MACHINE LEARNING BASED TECHNIQUES
Anomaly detection techniques can be sub categorized into Statistical Approaches, Cognition and Machine learning. Today, Machine learning techniques are popular for so many real time problems. Machine learning techniques are based on explicit or implicit model that enables the patterns analyzed to be categorized. It can be categorized into Genetic Algorithms, Fuzzy Logic, Neural Networks, Bayesian networks and outlier detection [3] [4].

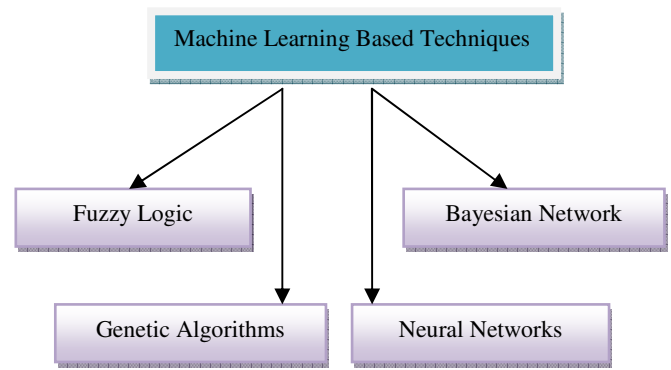

Figure 1: Categories of Machine learning based techniques

### 3.1 Fuzzy Logic
Fuzzy logic is derived from fuzzy set theory under which reasoning is approximate rather than precisely derived from classical predicate logic. Fuzzy techniques are thus used in the field of anomaly detection mainly because the features to be considered can be seen as fuzzy variables. Although fuzzy logic has proved to be effective, especially against port scans and probes, its main disadvantage is the high resource consumption involved. [5].

### 3.2 Genetic Algorithms

Genetic Algorithms are biologically inspired search heuristics that employs evolutionary algorithm techniques like crossover, inheritance, mutation, selection etc. So, genetic algorithms are capable of deriving classification rules and selecting optimal parameters for detection process. The application of Genetic Algorithm to the network data consist primarily of the following steps [6]:

i. The Intrusion Detection System collects the information about the traffic passing through a particular network.
ii. The Intrusion Detection System then applies Genetic Algorithms which is trained with the classification rules learned from the information collected from the network analysis done by the Intrusion Detection System.
iii. The Intrusion Detection System then uses the set of rules to classify the incoming traffic as anomalous or normal based on their pattern.

### 3.3 Neural Networks

A neural network is the ability to generalize from limited, noisy and data that is not complete. This generalization capability provides the potential to recognize unseen patterns, i.e., not exactly matched patterns that are different from the predefined structures of the previous input patterns. The neural network has been recognized as a promising technique for anomaly detection because the intrusion detector should ideally recognize not only previously observed attacks but also future unseen attacks [7].

### 3.4 Bayesian Networks

A Bayesian network is a model that encodes probabilistic relationships among the variables of interest. This technique is generally used for intrusion detection in combination with statistical schemes, a procedure that yields several advantages, including the capability of encoding interdependencies between variables and of predicting events, as well as the ability to incorporate both prior knowledge and data [5].

### 4. PROS/CONS OF ANOMALY DETECTION

Because anomaly based systems are capable of detecting misuse based on network and system behavior, the type of misuse does not need to be previously known. This allows for the detection of misuse a signature based system may not detect. While behavior on a system or a network can vary widely, anomaly based systems have the tendency to report a lot of false alarms. The art of effectively identifying "normal" activity vs. truly abnormal is extremely challenging [8] [3].

| Techniques | Pros/Cons |
|---|---|
| Fuzzy Logic | • Reasoning is Approximate rather than precise.<br>• Effective, especially against port scans and probes.<br>• High resource consumption involved. |
| Genetic Algorithm | • Biologically inspired and employs evolutionary algorithm.<br>• Uses the properties like Selection, Crossover, and Mutation.<br>• Capable of deriving classification rules and selecting optimal parameters. |
| Neural Network | • Ability to generalize from limited, noisy and incomplete data.<br>• Has potential to recognize future unseen patterns. |
| Bayesian Network | • Encodes probabilistic relationships among the variables of interest.<br>• Ability to incorporate both prior knowledge and data. |

**Table 1: Various Machine learning based anomaly detection Techniques**

### 5. CONCLUSION

In this review paper, types of intrusion detection have been discussed along with the brief introduction of the categories of the Anomaly detection which is one of the types of IDS. Machine learning based anomaly detection techniques are also discussed from the suitable references.

### 6. REFERENCES

[1] Karen Scarfone and Peter Mell, "Guide to Intrusion Detection and Prevention Systems (IDPS)," Department of commerce, National Institute of Standards and Technology, Gaithersburg, 2007.

[2] Asmaa Shaker ashoor and Sharad Gore, "Intrusion Detection System (IDS): Case Study," in *IACSIT Press*, Singapore, 2011, pp. 6-9.

[3] Chris Petersen. (2012, February) LogRhythm website. [Online]. www.logrhythm.com

[4] V. Jyothsna, V.V Ramaprasad, and K Munivara Prasad, "A Review of Anomaly based Intrusion," *International Journal of Computer Applications*, vol. 28, no. 7, pp. 26-35, August 2011.